\documentclass{onethree}

\usepackage{amsmath,amssymb}
\usepackage{array}
\usepackage{bm}
\usepackage{wrapfig}
\usepackage{enumitem}
\usepackage{colortbl}
\newenvironment{fixedfigure}{%
  \par\medskip\noindent\begin{minipage}{\linewidth}\captionsetup{type=figure}%
}{\end{minipage}\par\medskip}

\title{X-World 1.6: Building Controllable Virtual Cities with World Models}
\author{GWM Team, XPeng Inc.}
\affiliation[]{\href{https://xworld16.github.io}{~Project Page}}
\renewcommand{\teasercontent}{%
  \noindent\begin{minipage}{\textwidth}
    \centering
    \includegraphics[width=0.95\linewidth]{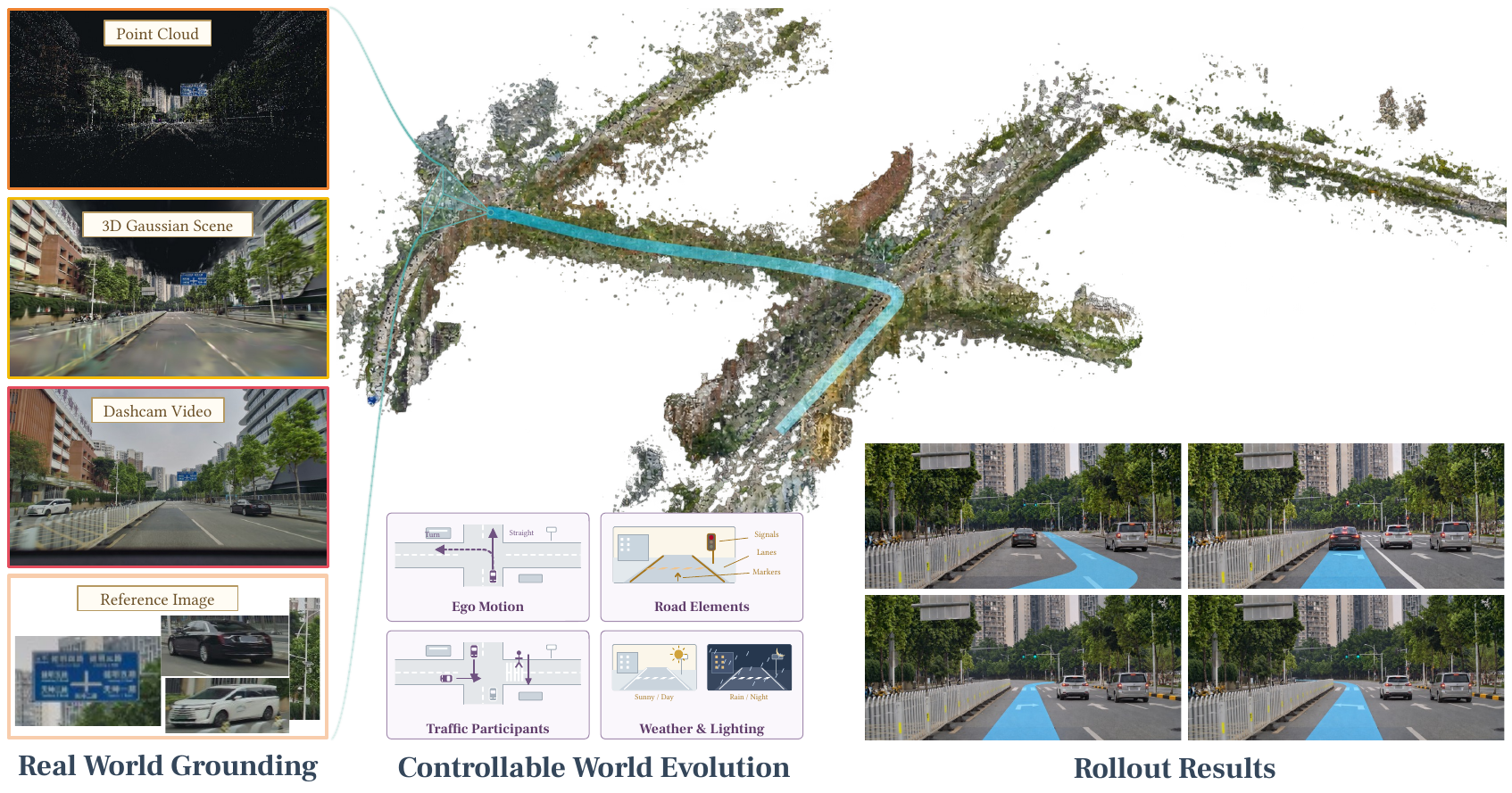}
    \captionof{figure}{From real-world scene references to controllable multi-camera generation.}
    \label{fig:teaser}
  \end{minipage}\par
  \vspace{4mm}
}

\hypersetup{
  pdftitle={X-World 1.6: Building Controllable Virtual Cities with World Models},
  pdfauthor={GWM Team, XPeng Inc.},
  pdfsubject={Technical report}
}

\abstract{
Deploying generative world models for end-to-end closed-loop evaluation and policy
learning requires precise scenario control, consistent observations,
and sustained generation quality during extended interactions.
This report presents \emph{X-World~1.6}, advancing X-World~1.0 toward a more
reliable and versatile driving simulator.
The upgraded base model offers finer-grained control over traffic
semantics, including traffic lights and road markings, together with
improved spatial consistency across cameras.
Virtual City capabilities broaden scenario construction through a
family of models conditioned on diverse 2D and 3D references, including
support for generating multi-camera future videos from a single image.
An enhanced post-training framework enables efficient causal, few-step
generation with stronger temporal coherence, better quality retention
over longer rollouts, and improved alignment with human preferences.
We also introduce a dedicated evaluation suite to support systematic
assessment and guide model iteration.
Together, these upgrades expand the scenarios X-World can simulate
and improve the efficiency and reliability of its generated observations,
supporting broader use in autonomous-driving development.
}

\begin{document}
\raggedbottom
\maketitle

\clearpage
\begingroup
% \small
\setlength{\parskip}{0pt}
\setlength{\cftbeforesecskip}{5pt}
\tableofcontents
\endgroup
\clearpage

\section{Introduction}
\label{sec:introduction}

X-World~1.0~\cite{xworld2026} established controllable, action-conditioned multi-camera generation as a practical component of autonomous-driving development~\cite{gaia12023,gaia22025,m4world2026,hu2026zyt}. Its streaming autoregressive interface already supported closed-loop simulation, online reinforcement learning, and synthetic data generation in XPENG's VLA~2.0 workflow. The starting point for X-World~1.6 is therefore an existing interactive simulator, with controls over ego motion, traffic participants, road geometry, and global appearance.

X-World~1.6 advances this foundation through stronger base-model controllability, more effective long-horizon post-training, and a more complete evaluation and iteration pipeline. Together, these developments support \textbf{Virtual City}: recorded real-world locations become configurable interactive environments for the established closed-loop evaluation and policy-training workflows.

\textbf{Stronger and broader scene control.}
We strengthen existing controls, add traffic-light positions and colors and road-surface markings, and improve geometric association across camera views. The principal extension beyond element-level control is \textbf{absolute static scene control}. Scene references from 3D Gaussian representations~\cite{gaussiansplatting2023,chen2026quantifying,zhang2024gaussian}, point clouds, and visual observations constrain persistent geometry and appearance~\cite{chen2026vggt} while ego trajectories and supported traffic conditions change. This provides evidence for buildings, roadside structures, and location-specific details that sparse road annotations do not determine. The objective is to preserve a specified place within the available reference coverage, while allowing scenario-dependent content to evolve. These conditioning paths use independently trained models; the common interface does not imply a single model accepting every reference modality.

\textbf{More effective long-horizon post-training.}
The two-stage training organization is retained: a bidirectional base model learns controllable generation, and post-training develops causal, few-step interaction~\cite{selfforcing2025,causalforcingpp2026}. Connected Self Forcing (CSF) restores feedback from later predictions through the history on which they depend~\cite{csf2026}. Longer teacher supervision and student rollouts extend this optimization to longer temporal contexts. Reward-guided refinement further targets visual quality and ego-motion adherence. The emphasis is on maintaining useful observations during continued interaction, where errors from earlier predictions can affect subsequent generation.

\textbf{Evaluation as part of model iteration.}
The human annotation pipeline provides a shared taxonomy and localized annotations of generation failures. VLM-assisted pre-labeling and human correction make this process more efficient, while automated benchmarks compare temporal and cross-view consistency across versions and training variants. We report both the annotation-efficiency benefit and the model comparisons. Scene-reference fidelity, internal consistency, and downstream behavior reproduction are kept distinct so that each result supports the capability it actually measures. Together with the model and training changes, this evaluation infrastructure supports the improvements in existing closed-loop applications reported in this work.

\section{Data}
\label{sec:data}

X-World~1.6 expands and curates real-world driving recordings to support controllable multi-camera generation and interaction-rich training. As illustated in Figure.~\ref{fig:data} (a), the data pipeline combines capture-quality screening, scenario tagging, and targeted selection for supervised fine-tuning (SFT). 
%Location-specific scene references and their conditioning interfaces are described separately in Section~\ref{sec:method}.

\subsection{Data Scaling}
\label{sec:data_scaling}

Compared with X-World~1.0~\cite{xworld2026}, the candidate recording pool is four times as large and the selected training dataset is twice as large. The distinction between candidate and training data matters: the larger pool allows stricter selection while retaining more usable clips and a wider range of driving situations. Both versions construct their training datasets by selecting clips from the available recordings.

\subsection{Quality Screening}
\label{sec:data_cleaning}

We exclude video segments affected by capture defects, including abrupt luminance flicker, horizontal color stripes, lens contamination or obstruction, strong color casts, and overexposure. Detection thresholds are calibrated through manual review of flagged samples and samples close to each threshold. Screening degraded camera streams uses both absolute image-quality signals and comparisons with other cameras. This helps distinguish a faulty view from environmental conditions such as fog, rain, or low light that can affect several cameras together. Candidate screening is followed by VLM verification.

\subsection{Scenario Tags}
\label{sec:tag_expansion}

The annotation taxonomy now contains more than 150 tags, with additions covering data quality, fine-grained agent behavior, and corner cases. Quality tags identify potential capture issues, finer behavior labels distinguish maneuvers previously grouped together, and corner-case tags make uncommon situations retrievable. These tags support targeted retrieval and control of the selected scenario mix. We also refine the tagging algorithms to reduce incorrect and missed labels. The same annotations provide the behavioral context used in SFT selection.

\subsection{SFT Construction and Distribution}
\label{sec:sft_data}

Routine lane keeping and steady cruising dominate the natural recording distribution. We increase the representation of less common maneuvers through two complementary selection strategies. Kinematics-based selection identifies sustained high-curvature motion and deceleration to a complete stop directly from ego trajectories. Behavior-based selection uses tags to identify contextual events such as intersection turns, merges, and overtaking.

\begin{figure}[t]
    \centering
    \includegraphics[width=\linewidth]{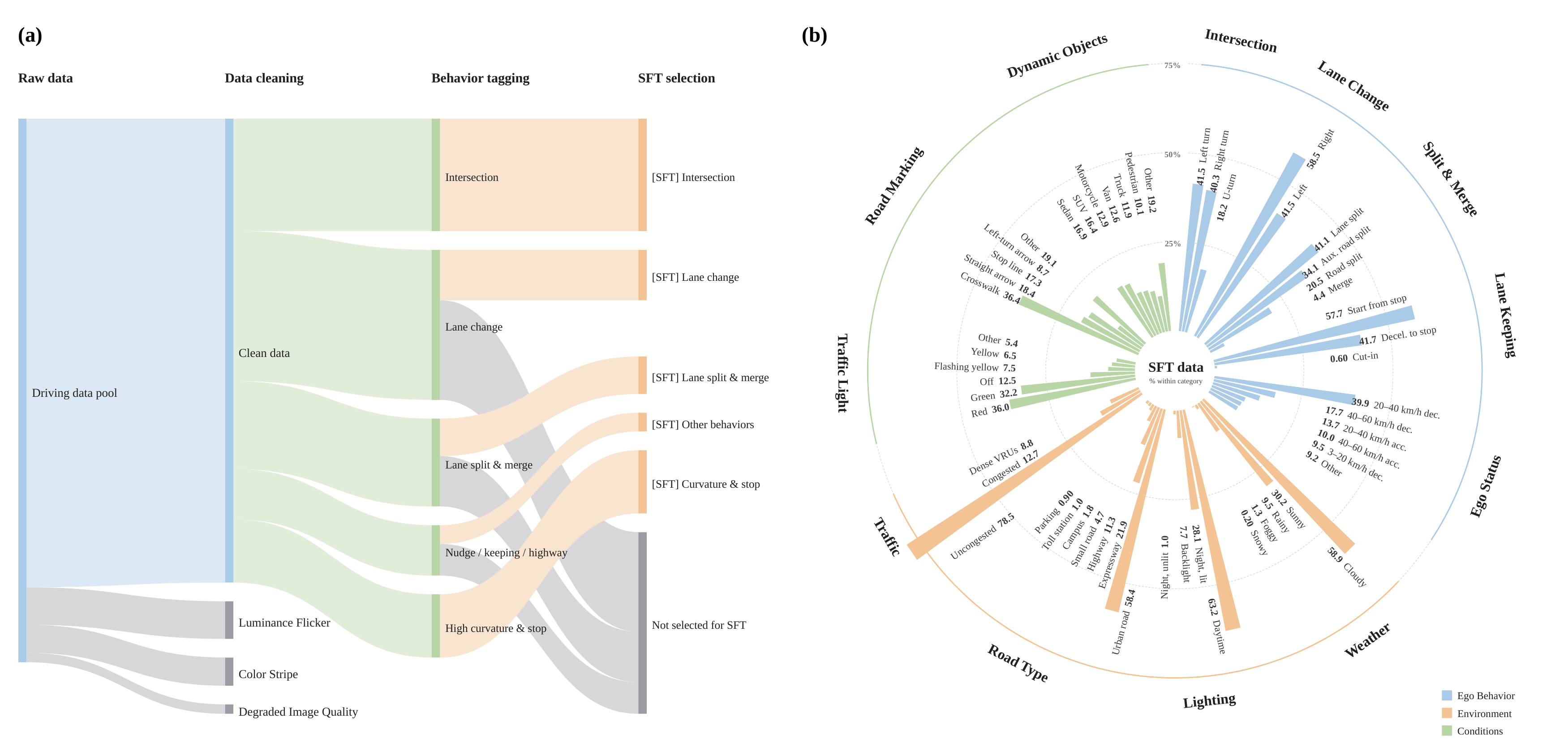}
\caption{(a) Data curation through quality filtering, behavior tagging, and scenario selection; flow widths are illustrative. (b) Tag distribution across ego behavior, environment, and conditions; bar lengths indicate within-category percentages.}
    \label{fig:data}
\end{figure}

A clip may contain several occurrences of a behavior, so each tag instance serves as a sampling unit. Category-specific resampling ratios favor scenarios with richer interactions. Combining motion-based selection with behavior tags retains both the executed maneuver and its driving context, without assuming that one is a substitute for the other. Figure~\ref{fig:data} (b) organizes selected SFT tag distributions into three families: ego behavior, environment, and conditions. Ego behavior covers intersection maneuvers, lane changes, split and merge events, lane keeping, and ego status; environment covers traffic, road type, lighting, and weather; conditions cover dynamic objects, road markings, and traffic lights. Percentages are normalized within each labeled category. They describe the selected SFT data, not the full training corpus or real-world occurrence frequencies.

\textbf{Behavior distribution.} Within the Intersection category, left turns, right turns, and U-turns account for 41.5\%, 40.3\%, and 18.2\%, respectively. Other behavior panels describe lane transitions and changes in ego motion, including starting, stopping, and acceleration or deceleration. These breakdowns complement the two selection mechanisms: trajectories identify the motion, while behavior tags provide its event context.

\textbf{Environmental context.} Urban roads account for 58.4\% within Road Type, while rainy conditions account for 9.5\% within Weather. The environment panels also distinguish traffic conditions and lighting, showing the context accompanying selected maneuvers.

\textbf{Scene conditions.} Red and green signals account for 36.0\% and 32.2\% within Traffic Light. The Road Marking panel covers crosswalks, directional arrows, and stop lines; Dynamic Objects distinguishes vehicle classes and pedestrians. These distributions describe the elements represented by the selected condition tags. Together, the three families connect SFT sampling to actions, their environmental context, and condition following. The signal and marking categories are particularly relevant to the traffic-light and road-surface-marking control interfaces added in X-World 1.6.

\section{Method: Controllable Base Model}
\label{sec:method}
X-World 1.6 strengthens existing controls over ego motion, traffic participants, road geometry, and appearance; adds traffic-light and road-marker conditions; and introduces absolute static scene control. The base-model phase learns these conditioning interfaces through bidirectional video training. Subsequent post-training improves their use during sustained autoregressive interaction.

\subsection{Task and Architecture}
\label{sec:city_formulation}
The model generates synchronized seven-camera video $\mathcal Y$ from scenario controls $\mathcal C$, target camera configurations $\Pi$, and, where available, location-specific scene evidence $\mathcal S$:
\begin{equation}
\mathcal Y\sim p_\theta(\,\cdot\mid\mathcal S,\mathcal C,\Pi).
\label{eq:virtualcity}
\end{equation}
Here $\mathcal C$ contains ego motion, traffic participants, road geometry, traffic-light states, road markers, and environmental text. Camera poses must agree with the requested ego motion. The distinction between scene evidence and controls allows the same location to support different driving scenarios: persistent structures remain constrained by the reference, while supported traffic and appearance conditions can change.

\begin{figure}[htbp]
\centering
\includegraphics[width=\linewidth]{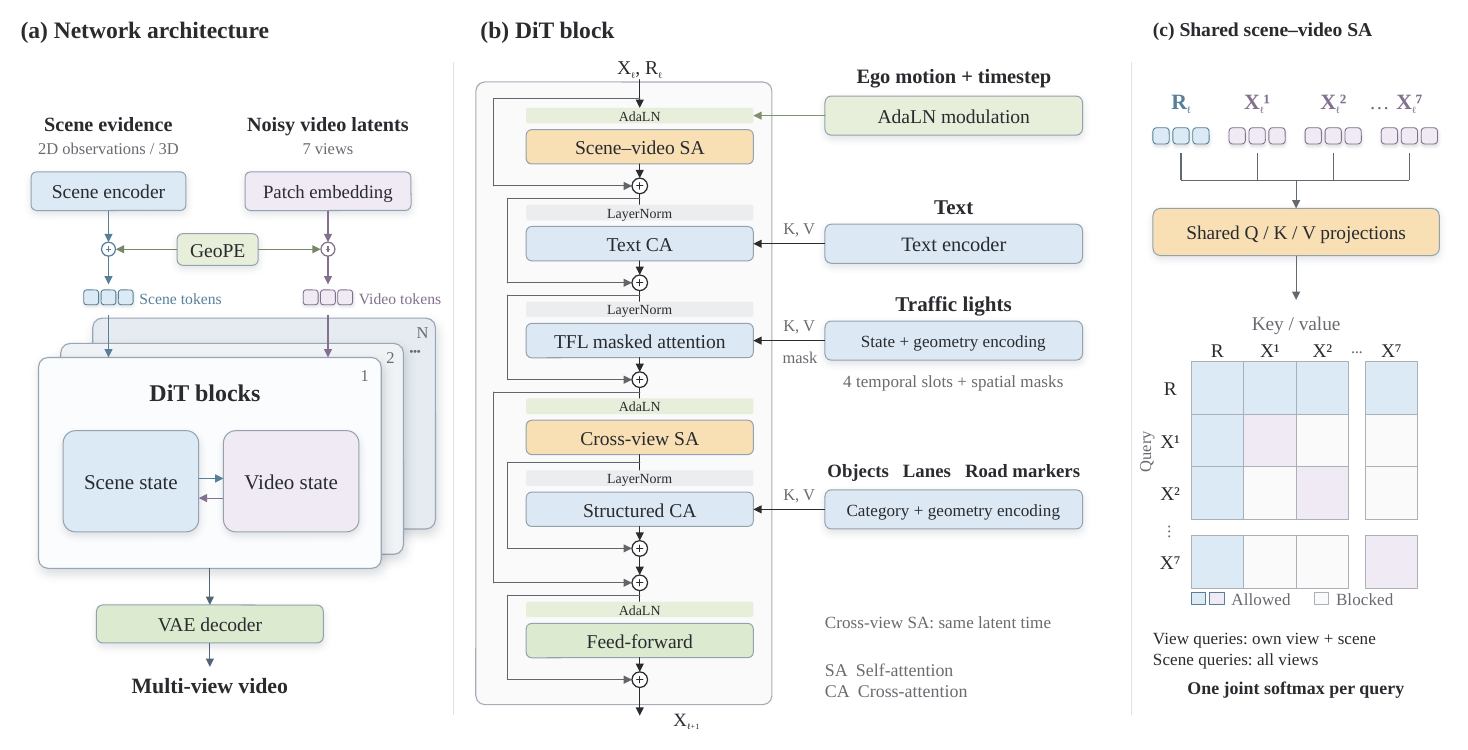}
\caption{\textbf{Base-model conditioning paths.} Separate models use visual references or 3D scene representations to guide seven-camera generation. Dashcam conditioning is a related synchronized-completion capability.}
\label{fig:virtual_city_architecture}
\end{figure}

Figure~\ref{fig:virtual_city_architecture} shows the conditioning paths around the multi-camera video backbone. All model variants share the same backbone architecture and are fine-tuned separately for their respective conditioning modalities and tasks. Scene evidence is encoded as geometry-aware scene tokens and fused with target-video tokens through joint self-attention; temporal and cross-view attention model the generated video. The model family also supports initialization from a single image and synchronized dashcam-to-surround completion. The latter follows the source video's recorded evolution and does not provide independently editable future dynamics.

\textbf{Shared geometric representation.}
Geometry-aware positional encoding (GeoPE) associates image patches with viewing rays in a common scene coordinate system~\cite{li2025cameras,wu2026rayrope}. Each descriptor combines ray direction, a symlog-scaled Pl\"ucker moment, and horizontal and vertical angular extent. It relates reference observations to target viewpoints without requiring surface depth. To account for temporal compression by the WAN2.2 VAE~\cite{wan2025}, GeoPE constructs a geometry embedding for each latent from the ordered geometric descriptors of its corresponding frames: a still image or the first video latent has one-frame support, while subsequent latents have four-frame support. A shared encoder adds these geometry embeddings to reference and target tokens. Figure~\ref{fig:geope} in Appendix~\ref{app:method} illustrates the spatial descriptor, ordered temporal support, and geometry--visual fusion; the corresponding definitions and masking details accompany the diagram.

\subsection{Enhanced Basic Control}
\label{sec:basic_control}
Ego actions retain the velocity, curvature, roll, and pitch representation of X-World 1.0~\cite{xworld2026}, encoded with symlog normalization, Fourier features, and an MLP before action-dependent modulation. Dynamic participants combine semantic features with spatial states; lanes and road boundaries describe continuous road geometry. Environmental text continues to control weather, illumination, and appearance.

The action embedding is combined with the diffusion-timestep embedding to parameterize AdaLN-Zero. Dynamic-object category features originate from the text encoder and are combined with Fourier-encoded coordinates. These inherited interfaces remain distinct from absolute static control: specifying an object or lane constrains a selected element, whereas a scene reference supplies the surrounding environment. A traffic light is static in position even though its state changes over time.

\textbf{Road markers and shared structured conditioning.}
X-World 1.6 adds discrete road-surface markings represented by category features, sampled geometry, and adjacent-point offsets. Dynamic objects, road markers, and lanes use separate encoders, preserving the representation appropriate to each condition type while mapping their outputs to the same DiT width. Their tokens and validity masks are concatenated for a shared cross-attention layer after cross-view fusion. In 1.0, dynamic objects used a per-view attention path and lanes a separate post-fusion branch. The shared branch reduces repeated projections while exposing the fused camera features to all three condition groups. Text retains its own per-view branch.

\textbf{Traffic-light conditioning.}
Signals require both spatial localization and source-frame timing. Their visible support is small and camera-dependent, and their color may change within the four source frames compressed into one video latent. Ordinary condition-token validity masks do not localize updates to a particular signal; temporal averaging can also mix distinct light states. We therefore use a dedicated branch with per-source-frame state tokens and per-light spatial masks.

 The geometry of the traffic light condition is projected into each calibrated camera and pooled onto the latent grid. For spatial token $p$, light $n$, view $v$, latent step $\ell$, and source-frame slot $f$, the mask is
\begin{equation}
M^{\mathrm{TFL}}_{v,\ell,f}(p,n)
=\mathbf 1[n\text{ is valid}]
\mathbf 1[p\in\Omega_{v,\ell,f,n}],
\label{eq:many_futures_tfl_mask}
\end{equation}
where $\Omega_{v,\ell,f,n}$ is the projected signal region. The branch sits after per-view text attention and before cross-view fusion. Its four head groups query the corresponding source-frame states, retaining changes that would be lost by averaging. Tokens outside valid signal support receive no direct update from this branch. Appendix~\ref{app:method} records the first-latent convention.

Static-scene references and these controls play complementary roles. Changing traffic does not require replacing the location asset, but editing lane or marker geometry may require compatible changes to that asset. Traffic-light schedules and actor trajectories are specified separately and must agree when constructing coordinated scenarios. Environmental edits preserve location identity through geometry and distinctive structures, while allowing condition-dependent shadows and reflections to change.

\subsection{Absolute Static Scene Control}
\label{sec:absolute_static_control}
Sparse road annotations do not fully specify the buildings, guardrails, and surface details of a particular place. Absolute static scene control supplies persistent evidence of that environment throughout generation. Point clouds, 3D Gaussian Splatting (3DGS)~\cite{gaussiansplatting2023} assets, or reference observations from vehicle logs provide the scene evidence; the model associates it with the changing target viewpoints while retaining independent action and traffic controls. Each modality uses a separately trained model with scene conditioning injected through joint self-attention. Content outside the available coverage is completed using the learned prior.

\subsubsection{Point-Cloud Conditioning}
\label{sec:pointcloud_world_generation}
We use a static point cloud as a 3D record of the scene, with color where available. For each of the seven target cameras, we render an RGB image and a depth map at each requested pose. RGB provides a rough view of the scene, while depth gives the distance to the visible points. Zero depth marks areas without point-cloud observations, which the model can fill in during generation.

The RGB and depth sequences are VAE-encoded and converted into scene tokens. Depth preprocessing keeps zero as the missing-data marker. GeoPE adds camera geometry to the scene and target-video tokens, helping the model connect the rendered views to the video being generated. The scene tokens are concatenated with target-video tokens and fused through joint self-attention in the DiT blocks.

During training, we render the point cloud along the recorded camera path and use the real seven-camera video as the denoising target. At inference, the requested ego trajectory determines both the rendering poses and the action condition. The same point cloud can therefore support different driving paths and traffic conditions while providing a common static scene reference.

\subsubsection{3DGS References and Trajectory Decoupling}
We use a 3DGS scene to render reference videos of the static environment. Moving objects are removed from these references in one of two ways: we model them separately following OmniRe~\cite{chen2025omnire} and hide them during rendering, or mask them out when reconstructing the scene. Both approaches use SAM3 segmentation; the first also requires object tracking. Masking provides an alternative when tracking is unreliable. This prevents vehicles in the original recording from becoming part of the fixed background.

The rendered reference videos are converted into scene tokens and augmented with GeoPE. These tokens are concatenated with target-video tokens and fused through joint self-attention in the DiT blocks. They provide the scene's appearance and geometry, while action and traffic controls guide the generated motion.

To teach the model to use reference videos from a different camera path, we render them along alternative trajectories through the reconstructed scene. The real recorded video remains the denoising target, and its original ego trajectory remains the action condition. We construct approximately one million such reference--target pairs offline. This lets the model learn scene details from the reference while following the target trajectory.

We also encode the relative position and orientation of the reference and target cameras. This information helps the model use the reference scene from the requested viewpoints. Appendix~\ref{app:pose_signal} gives the pose definition and encoding details.

\begin{figure}[t]
\centering
\includegraphics[width=0.96\linewidth]{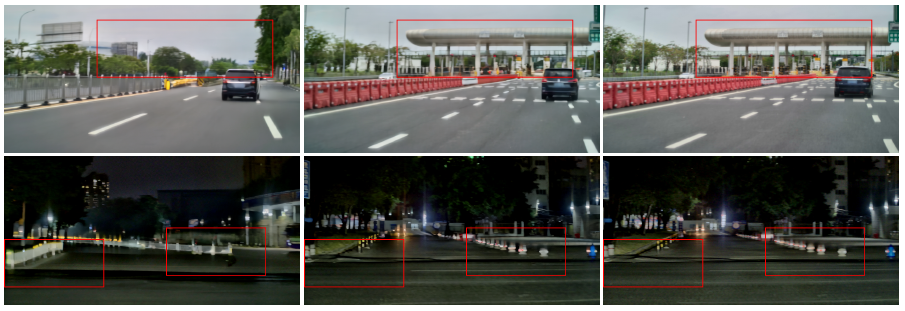}
\caption{Generation without and with 3DGS absolute static scene (ASC) conditioning, alongside ground truth. Each row shows a different scene.}
\label{fig:asc_qualitative}
\end{figure}

At inference, the rendered reference guides the static surroundings while the model follows the requested ego motion and traffic conditions. Figure~\ref{fig:asc_qualitative} compares generated videos with and without this reference against the recorded scene.

\subsubsection{Reference-Image Conditioning}
\label{sec:r2v}
We use reference image patches to guide multi-camera video generation. Each reference image is independently VAE-encoded and converted into patch tokens that represent local appearance, such as building facades and roadside structures. GeoPE adds information from the camera calibration and pose, helping the model relate reference patches to the target views. This provides scene information directly from images without requiring explicit 3D reconstruction.

The reference tokens are concatenated with target-video tokens and fused through joint self-attention in the DiT blocks. All seven cameras share the same reference tokens, so each view can use relevant scene details during generation. The reference images provide scene appearance, while ego actions and traffic controls guide how the generated scene evolves.

During training, we sample multiple reference images and their camera geometry from the same recorded scene and use the real multi-camera video as the denoising target. At inference, the reference images remain fixed while the model generates video along the requested camera path, helping preserve the location's appearance as viewpoints and traffic conditions change.
\subsection{Bidirectional Base-Model Training}
\label{sec:training_setup}
We initialize the base models from Wan2.2-TI2V-5B~\cite{wan2025} and train separate variants for the conditioning routes described above. Training uses synchronized seven-camera videos and updates the DiT while keeping the VAE and text encoder frozen. Bidirectional temporal attention lets each frame use information from both earlier and later frames in the training clip.

Depending on the variant, inputs can include initial images, reference images or videos, or rendered 3D scene views, together with action, text, camera geometry, and scene controls. The real multi-camera video provides the training target in every case.

\textbf{Denoising objective.} We use Rectified Flow~\cite{liu2022flow} to train the model to denoise video latents. Given clean latents $x_0$ and Gaussian noise $\epsilon\sim\mathcal N(0,I)$, we form $x_\tau=(1-\sigma_\tau)x_0+\sigma_\tau\epsilon$. All seven views use the same noise level, but their noise is sampled independently. The model predicts the flow velocity from the noisy video and the available conditions:
\begin{equation}
\mathcal L_{\mathrm{FM}}
=\mathbb E\!\left[w(\tau)\,
\operatorname{MSE}\!\left(v_\theta(x_\tau,\tau;\tilde c),\epsilon-x_0\right)\right].
\label{eq:base_flow}
\end{equation}
Here, $\tilde c$ denotes the conditions remaining after random condition dropout, which removes selected inputs during training. The loss averages prediction errors over latent elements in all seven views, and $w(\tau)$ weights the different noise levels.

\textbf{Training schedule.} We first use a higher condition-dropout rate to train with incomplete combinations of inputs. When all droppable conditions are absent, we keep randomly selected temporal latent positions clean so they act as observed frames for video completion. These clean positions use a zero timestep and a zero velocity target. This is the temporal latent inpainting used in this step.
We then continue from this checkpoint with a lower condition-dropout rate, placing more emphasis on following the supplied conditions. Temporal inpainting is disabled in this step; the model architecture and training objective stay the same.
Both steps belong to base-model training. The resulting bidirectional checkpoints initialize the causal models and serve as real-score teachers for distribution matching distillation in the post-training stage described in Section~\ref{sec:posttraining}. 
\section{Post-training for Long-Horizon Interaction}
\label{sec:posttraining}

\begin{figure}[t]
    \centering
    \includegraphics[width=\linewidth]{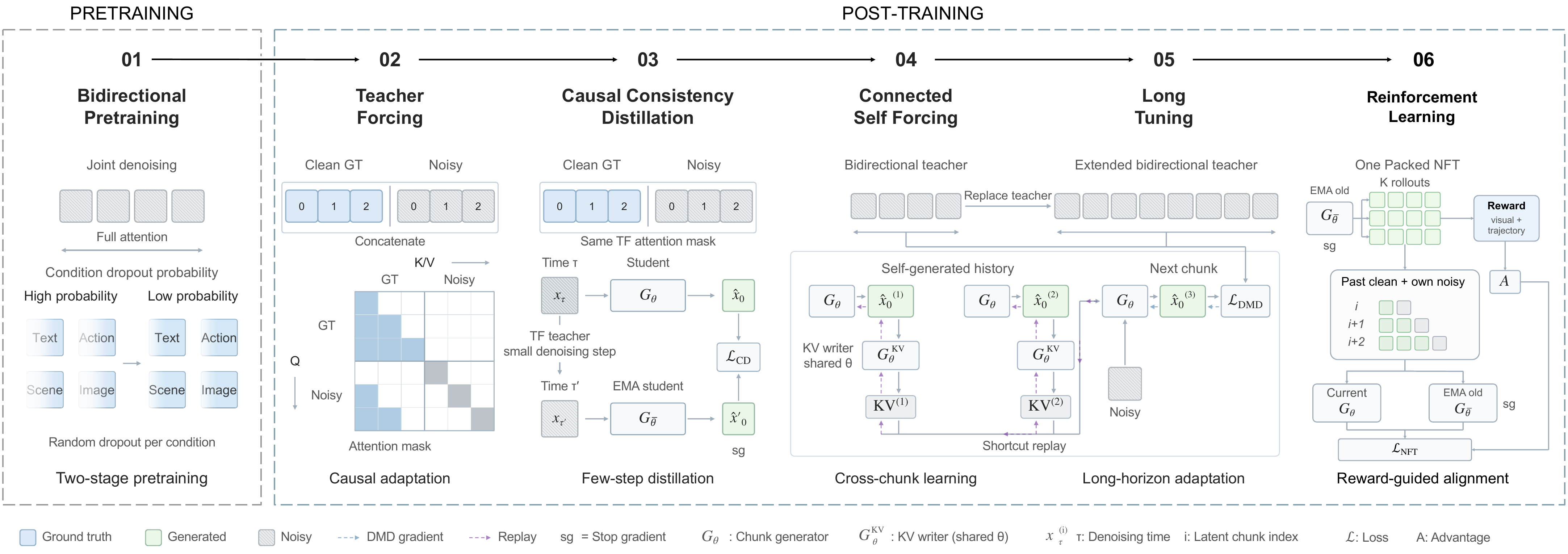}
    \caption{\textbf{Two-stage training pipeline.} Block 01 develops the controllable base model. Blocks 02--06 adapt it for causal few-step generation, restore cross-chunk feedback, extend the training horizon, and apply reward-guided refinement.}
    \label{fig:training_pipeline}
\end{figure}

X-World~1.0 already supported causal, few-step, long-horizon generation. X-World~1.6 improves the quality and consistency of extended interaction through a more effective post-training pipeline. Starting from the controllable bidirectional base model (Section~\ref{sec:training_setup}), we establish causal few-step prediction, train on generated histories with our Connected Self Forcing (CSF)~\cite{csf2026}, extend teacher and student horizons, and refine visual and motion adherence with rewards. Figure~\ref{fig:training_pipeline} shows how these steps connect.

\subsection{Causal Few-Step Adaptation}
\label{sec:teacher_forcing}
\label{sec:causal_consistency_distillation}

Teacher forcing initializes from the final bidirectional checkpoint and replaces full temporal attention with a block-causal mask, preserving bidirectional attention within each chunk. Clean ground-truth (GT) latents and their noisy counterparts are concatenated for parallel training: each noisy chunk sees preceding clean chunks and its own noisy tokens, but neither its clean target nor future chunks. The clean-history branch is also causally masked. Training retains the base model's flow-matching objective, with supervision only on the noisy branch. Noise timesteps are sampled independently per chunk and shared across its seven views; the flow-matching loss retains timestep weighting.

We then apply Causal Consistency Distillation, following Causal Forcing++~\cite{causalforcingpp2026,consistencymodels2023}. A frozen TF teacher takes an Euler step with classifier-free guidance to construct adjacent noise states. The student and its exponential moving average (EMA) counterpart predict a consistent clean endpoint from these states under the same GT history. Noise-state pairs are shared across synchronized views and chunks. The distilled EMA weights initialize CSF, stabilizing the training on few-step self-generated context.

\subsection{Connected Self Forcing}
\label{sec:connected_self_forcing}

At deployment, generated observations become the context for subsequent predictions. CSF~\cite{csf2026} trains directly on this autoregressive distribution, building on Self Forcing~\cite{selfforcing2025}. The generator produces synchronized seven-view chunks through few-step denoising, conditioned on scene inputs, controls, and earlier generated chunks. A clean KV-writing pass $G_\theta^{\mathrm{KV}}$, sharing the generator's parameters, then encodes each generated chunk into the historical cache. This separates the prediction of a chunk from its use as context for later predictions.

Distribution Matching Distillation (DMD)~\cite{dmd2023,dmd22024} provides direct supervision: a frozen bidirectional real-score teacher supplies the target distribution, while a trainable fake-score model tracks the generator. Evaluating re-noised generated latents with both models yields gradients on the generated chunks.

The additional challenge is propagating feedback through history. Standard Self Forcing detaches historical KV caches to save memory, so later losses cannot reach the computations that produced earlier context. A historical chunk receives direct output supervision, but no feedback on how its cached representation affects subsequent predictions. CSF restores this connection through \textbf{Shortcut Gradient Replay}.

During the initial autoregressive rollout, the graph is detached while retaining the states needed for replay. Given a later chunk's DMD gradient, CSF replays its denoising computation with selected historical KV states made differentiable. It then propagates the recovered cache gradients through the corresponding KV-writing computations to the historical generated chunks, and through their replayed denoising computations to the shared generator parameters. A historical chunk $j$ consequently receives both its direct DMD gradient and cross-chunk feedback, $\mathbf g_{\mathrm{DMD}}^{(j)}+\mathbf g_{\mathrm{cross}}^{(j)}$.

Replay avoids retaining the full autoregressive computation graph while optimizing earlier context for its own quality and its usefulness to future predictions. It is a training procedure: causal attention, few-step denoising, and clean-cache writing remain unchanged at inference. The consistency comparisons in Section~\ref{sec:consistency_results} examine the resulting trade-offs against ordinary Self Forcing.

\subsection{Long-Horizon Tuning}
\label{sec:long_horizon_tuning}

Cross-chunk feedback operates within the temporal range covered by training. To improve consistency over longer horizons, we extend both the student rollout length and the teacher supervision window. This differs from Self-Forcing++~\cite{selfforcingpp2026} and LongLive~\cite{longlive2025}, which perform long-video tuning while retaining short-window teacher supervision. 
Specifically, we first fine-tune the bidirectional teacher on synchronized seven-view sequences that are three times as long as those used in its original training, while retaining its attention structure and flow-matching objective. We then continue CSF from the existing generator checkpoint, replacing the real-score teacher with this extended teacher and increasing both the student rollout length and the teacher supervision window.

The two changes are complementary: longer student rollouts expose the generator to its own accumulated history, while the longer teacher window provides distributional supervision over extended scene evolution. Few-step denoising, DMD, and Shortcut Gradient Replay are retained. This stage extends training without changing the causal architecture or inference procedure; Section~\ref{sec:consistency_results} compares SF and CSF with and without long-horizon tuning.

\subsection{Efficient RL for Causal Models}
\label{sec:diffusion_rl}

Reward-guided refinement supplements teacher-based supervision with visual preference and ego-motion adherence. Here reinforcement learning updates the world model itself; training a driving policy inside the resulting simulator is a separate application.

\textbf{One Packed NFT.}
\label{sec:one_packed_nft}
We build on DiffusionNFT~\cite{zheng2026diffusionnft}, which learns
from forward-noised generated samples without backpropagating
through sampling. For a causal video model, target prediction
must preserve the clean history available during generation.
Jointly noising all blocks under a standard causal mask instead
exposes each target to preceding noisy blocks, changing its
conditioning. A sequential implementation preserves clean
history by alternating target prediction with clean-cache
extension, requiring $H$ target-prediction calls for $H$ target
blocks per model role and candidate batch.
Our \emph{One Packed NFT} consolidates these target evaluations through
a teacher-forcing-style computation, using model-generated clean
history instead of ground-truth history. A frozen Old policy
generates a shared prefix and candidate suffixes, which remain
fixed with their rewards during inner updates. We separate
clean-history encoding from noisy-target prediction.
Current, Old, and Reference each construct detached, per-layer
KV banks using their respective parameters. Clean encoding
preserves block causality at every layer, preventing future
information from entering historical K/V. A structured mask
then lets each noisy target access its permitted clean history
and its own noisy tokens, enabling one packed target-prediction
call per model role and candidate batch.

\textbf{Joint Visual-Motion Reward}.
\label{sec:joint_reward_nft}
HPSv3~\cite{ma2025hpsv3} scores visual preference on frozen trajectories; a fixed inverse dynamics model (IDM)~\cite{agarwal2026cosmos} measures agreement between inferred and prescribed ego motion. Visual scores are averaged over frames associated with each temporal latent and camera. Each reward is standardized across valid candidates at each latent and clipped to $[-C,C]$, producing advantages $A_H$ and $A_I$. Visual normalization is camera-specific, while motion advantages are broadcast across views.

Although we focus on improving the IDM score, our experiments show that optimizing the IDM reward alone can lead to reward hacking. We therefore combine it with HPSv3 to stabilize training. We combine them as $A_{\mathrm{mix}}=\alpha_H A_H+\alpha_I A_I$, with nonnegative weights summing to one, and use the resulting preferences in the DiffusionNFT objective. A frozen-reference penalty limits deviation from the reference generator. This retains the standard NFT loss while adapting its rewards and causal evaluation to the multi-camera world model. In our experiment, we found that applying IDM reward may cause reward hacking but using it together with HPSv3 stables the training.

\section{Inference}
\label{sec:model_inference}

We deploy the four-step causal generator as a stateful service on T-Head PPU M810E accelerators. At each closed-loop step, updated ego motion and scene controls produce synchronized seven-camera observations conditioned on the generated history. The runtime preserves this interaction while reducing repeated computation and serving multiple rollouts concurrently. We report aggregate throughput across active sessions and per-session end-to-end latency separately. Unless otherwise stated, we conduct the following experiments without absolute static scene control.

\subsection{Stateful Interactive Execution}
\label{sec:inference_state}

Each session maintains its own self-attention KV cache, sampling-noise state, and causal VAE boundary state. Initial observations seed the history once. Subsequent requests update controls and denoise the active two-latent, eight-frame causal block against that history.

The chunk interface returns a full block; the per-frame interface accepts updated controls throughout the block's temporal span. Each per-frame request re-denoises the active block and overwrites its intermediate KV slots. After all eight frames are emitted, the clean-context pass $G_\theta^{\mathrm{KV}}$ commits the finalized keys and values before the window advances. Intermediate noisy predictions therefore never become history for later blocks.

The cache retains a fixed attention-sink prefix and a sliding window over recent history and the active block~\cite{xiao2024streamingllm}, keeping memory bounded as rollouts grow. Independent session caches and execution streams preserve state under interleaved requests; each session remains bound to one denoiser throughout its rollout.

\subsection{Efficient Execution}
\label{sec:inference_residual_reuse}

Within each session, we cache text features and their attention projections while the text condition remains unchanged. Action and scene-control encodings and projections that do not depend on the noisy latents or diffusion timestep are computed once per request and reused across denoising steps.

X-Cache~\cite{zeng2026xcache} reuses DiT residuals across related requests when feature and control similarities pass an adaptive gate. The first denoising step, first Transformer layer, and clean-KV commit always execute fully. The measured block-reuse rate is $45.22\%$ across 12 concurrent closed-loop sessions.

Fused DiT kernels, channels-last VAE execution, and AIU convolutions reduce kernel-launch and data-movement overhead. Causal VAE boundary snapshots avoid repeatedly decoding the same prefix. Appendix~\ref{app:systems} records selected implementation details relevant to these optimizations.

\subsection{Numerical Fidelity}
\label{sec:inference_exact}

Autoregressive feedback can amplify small numerical differences. The kernel and layout optimizations preserve the reference computation bit-exact across multi-session closed-loop rollouts; cross-chunk residual reuse (X-Cache) is the only approximation. Over $1344$ generated frames, the complete runtime achieves a mean frame PSNR of 53.20~dB and a minimum frame PSNR of 49.12~dB against the reference output. These comparisons measure numerical fidelity to the reference implementation, rather than visual fidelity to real driving footage.

\begin{table}[t]
    \centering
    \small
    \setlength{\tabcolsep}{5pt}
    \begin{tabular*}{\linewidth}{@{\extracolsep{\fill}}>{\raggedright\arraybackslash}p{5.4cm}cccccccc@{}}
        \toprule
        & \multicolumn{2}{c}{Obs/s $\uparrow$} & \multicolumn{2}{c}{Denoiser (s) $\downarrow$} & \multicolumn{3}{c}{Step latency (s) $\downarrow$} & \\
        \cmidrule(lr){2-3}\cmidrule(lr){4-5}\cmidrule(lr){6-8}
        Configuration & PPU & Node & P50 & Mean & P50 & Mean & P99 & Speedup $\uparrow$ \\
        \midrule
        Reference kernels & 0.156 & 0.625 & 5.96 & 5.92 & 19.04 & 18.68 & 23.80 & 1.00$\times$ \\
        + Channels-last VAE with fused causal input & 0.201 & 0.803 & 4.48 & 4.55 & 14.70 & 14.46 & 18.94 & 1.28$\times$ \\
        + Fused DiT kernels & 0.218 & 0.873 & 3.95 & 4.16 & 13.42 & 13.25 & 18.36 & 1.40$\times$ \\
        + AIU convolutions & \underline{0.237} & \underline{0.949} & \underline{3.69} & \underline{3.80} & \underline{12.49} & \underline{12.22} & \textbf{15.23} & \underline{1.52$\times$} \\
        + X-Cache & \textbf{0.320} & \textbf{1.281} & \textbf{2.10} & \textbf{2.62} & \textbf{8.69} & \textbf{8.94} & \underline{17.82} & \textbf{2.05}$\times$ \\
        \bottomrule
    \end{tabular*}
    \caption{Steady-state closed-loop serving throughput, server denoiser
    forward time, and per-session end-to-end step latency on a 4$\times$ PPU
    M810E node (12 concurrent sessions) with optimizations enabled
    cumulatively. Best and second-best values in each column are bold and
    underlined, respectively.}
    \label{tab:ppu_inference}
\end{table}

\subsection{Inference Performance}
\label{sec:inference_performance}

Table~\ref{tab:ppu_inference} uses a four-PPU M810E node with 12 concurrent sessions, BF16, and four denoising steps. Each seven-camera observation comprises two views at $512\times1024$, four at $416\times512$, and one at $448\times896$, totaling $2.30$~MPix. Each per-frame step includes condition preprocessing, DiT denoising, causal VAE execution, and frame delivery. Optimizations are enabled cumulatively under the same serving configuration; throughput is aggregated across sessions, while step latency is measured per session.

The complete configuration doubles throughput ($2.05\times$), reaching 0.320~obs/s/PPU. Eliminating VAE layout conversions and causal-padding copies accounts for most of the kernel/layout gain. X-Cache further reduces median denoiser time, while full clean-KV commit passes and latent-position transitions explain its higher P99 step latency relative to the configuration without X-Cache. The table retains both throughput and latency to expose this trade-off.

\section{Evaluation Framework and Model Assessment}
\label{sec:eval}

Our evaluation framework has three components: human evaluation, VLM-assisted evaluation, and an automated consistency benchmark. Human evaluation establishes the criteria for generation quality and provides reference judgments. The VLM learns this assessment standard to localize defects and reduce annotation effort. The automated benchmark provides repeatable measurements of temporal and cross-view consistency for comparing model versions and training variants. We first describe these components, then show how we apply them to annotation, evaluator validation, and model comparison.

Although VLM inference is automated, we distinguish it from the benchmark by what each evaluates. The VLM predicts human-defined anomaly categories and their spatial and temporal extent, supporting judgments of perceptual quality and plausibility. The benchmark measures specific consistency properties through fixed tracking, compression, and reconstruction procedures. These measurements reveal changes in appearance persistence, temporal predictability, and geometric agreement that per-camera anomaly judgments do not directly quantify. Conversely, consistency scores alone cannot establish overall perceptual quality or adherence to the requested scene and controls. We therefore use the two forms of evidence together, with human review providing the reference for anomaly-based assessment.

\subsection{Evaluation Components}
\label{sec:eval_components}

\subsubsection{Human Evaluation}
\label{sec:human_eval}

Human evaluation covers all seven camera views, with anomalies localized by image region and temporal interval. The assessment considers four dimensions: visual fidelity, scene consistency, dynamic plausibility, and condition adherence. These dimensions capture image degradation, the persistence and geometry of static structures, the identity and motion of surrounding agents, and consistency with controls such as object positions, lane geometry, road markings, and traffic-light states.

Annotators distinguish generation failures from legitimate occlusions and viewpoint changes. When multiple artifacts stem from the same underlying event, only the primary failure is recorded to avoid duplicate penalties. Video-level quality scores aggregate anomaly penalties weighted by camera priority, object size, and relevance to the driving region. A separate quality-control stage audits sampled annotations, with corrections required when disagreements materially affect scores or rankings.

The resulting annotations provide both video-level quality scores and spatiotemporal evidence of identified failures. They serve as training targets and held-out reference judgments for the VLM evaluator described next, linking automated anomaly detection to the same assessment standard.

\subsubsection{VLM-Assisted Evaluation}
\label{sec:vlm_eval}

Following CaC~\cite{wang2026cac}, we train a VLM to propose spatiotemporal anomalies before human review. For each camera video, it predicts whether an anomaly exists, its category, and its frame range and bounding boxes. Annotators accept, correct, or reject these predictions in the same format used by human evaluation. Corrected samples are then added to the training set of this evaluator, creating an iterative pre-labeling and verification workflow.

Starting from Qwen3-VL-8B-Instruct~\cite{bai2025qwen3vl}, we first fine-tune on single-frame anomaly recognition and localization, then on video clips for joint category, temporal, and spatial prediction. Training keeps normal and abnormal examples balanced and splits data by scene, preventing videos generated from one scene from appearing in both training and evaluation. GRPO~\cite{shao2024deepseekmath} further optimizes format validity, anomaly status, category agreement, and localization accuracy, with the largest weights on localization. Appendix~\ref{app:vlm_reward} gives the reward definition.

\subsubsection{Automated Consistency Benchmark}
\label{sec:consistency_eval}

The automated benchmark evaluates synchronized generated frames and their camera calibration, without requiring a paired real future or manual identity labels. It can therefore compare alternative generated futures under matched scene conditions. Three complementary modules measure object appearance, temporal predictability, and geometric reconstructability; their definitions and implementation settings are collected in Appendix~\ref{app:evaluation}.

\textbf{Object tracking.} SAM~3~\cite{carion2025sam3} and ByteTrack~\cite{zhang2022bytetrack} provide within-view object tracks, while FG-CLIP~2~\cite{xie2025fgclip2} encodes masked object crops. Mean appearance consistency measures similarity along a track; appearance decay measures how that similarity changes as the frame interval doubles. Higher consistency and a decay slope closer to zero indicate more persistent appearance. Calibrated rays through matched mask centers provide the mean cross-view ray residual, for which lower values indicate better geometric agreement.

\textbf{Temporal compression.} Each video is encoded with inter-frame prediction and with all-intra coding under fixed settings. Gains in decoded SSIM and PSNR, together with the inter/intra size ratio, measure how effectively recurring content supports temporal prediction. Greater quality gains and a lower size ratio indicate stronger predictability under this codec configuration.

\textbf{Reconstruction.} A fixed UniSplat~\cite{shi2026unisplat} model reconstructs the synchronized views with temporal memory and renders them at the supplied camera poses. SSIM, PSNR, and MAE against the generated inputs measure how well these observations can be explained by the shared reconstruction. Higher SSIM and PSNR and lower MAE indicate better reconstruction fidelity.

Tracking depends on valid associations, compression also reflects motion and content, and reconstruction depends on the fixed model's capacity and priors. We therefore retain all nine raw metrics to identify which consistency properties improve or regress. The average rank reported below summarizes these diagnostics within the compared model set.

\subsection{Applying the Evaluation Framework}
\label{sec:eval_application}

We use the human protocol and the VLM together to assess visible generation failures: the VLM proposes annotations, reviewers verify them, and corrected examples support subsequent evaluator training. Human annotations also provide the reference for testing whether the VLM preserves quality rankings and same-scene preferences. We use the automated benchmark for controlled comparisons of temporal and geometric consistency across model versions and training variants. The following experiments examine the annotation workflow, the evaluator's agreement with humans, and the model differences measured by the benchmark.

\subsubsection{Annotation Efficiency}
\label{sec:vlm_results}

On 150 generated videos, the same five annotators compare annotation from scratch with correction of VLM pre-labels. Pre-labeling reduces the total annotation time from 2 days to 1 day, a 50\% reduction; The final annotations remain subject to the human review and quality-control procedures described in Section~\ref{sec:human_eval}. This experiment measures the operational benefit of assistance; the next experiment tests the evaluator's agreement with human judgments directly.

\subsubsection{Agreement with Human Evaluation}
\label{sec:vlm_agreement}

We compare VLM predictions with human annotations on 30 driving scenes, each generated by X-World~1.0 and X-World~1.6, giving 60 videos and 420 camera views. For both sources of annotations, the human evaluation rubric converts defects into camera-view penalties, which are summed into a video-level total; lower totals indicate better quality. We measure correlation with human totals across videos and agreement on which model produces the better video within each scene. Human totals differ in 28 scenes; the remaining two tied scenes are excluded from pairwise accuracy. When human annotators prefer one video over the other, the VLM is counted as correct only if it prefers the same video; a tie is not counted as correct.

\begin{table}[t]
    \centering
    \small
    \setlength{\tabcolsep}{5pt}
    \begin{tabular}{lccc}
        \toprule
        Metric & Our evaluator & Qwen3-VL-8B (zero-shot) & Evaluation set \\
        \midrule
        Same-scene pairwise accuracy (\%) & 71.4 [53.6, 88.5] & 21.4 [7.4, 37.9] & 28 scenes \\
        Video-penalty Spearman $\rho$ & 0.671 [0.476, 0.815] & 0.097 [$-0.181$, 0.378] & 60 videos \\
        Video-penalty Pearson $r$ & 0.867 [0.540, 0.943] & $-0.074$ [$-0.238$, 0.207] & 60 videos \\
        \bottomrule
    \end{tabular}
    \caption{\textbf{Agreement with human evaluation on paired X-World~1.0 and X-World~1.6 generations.} Brackets give 95\% confidence intervals from a scene-level cluster bootstrap. The zero-shot baseline uses Qwen3-VL-8B-Instruct with the same inputs and prompts. The three metrics capture complementary aspects of agreement. Same-scene pairwise accuracy measures how often the VLM selects the video preferred by humans within a scene. Across all 60 videos, Spearman $\rho$ measures agreement in quality ordering using penalty ranks, while Pearson $r$ measures the strength of the linear relationship between the actual penalty values. Spearman therefore focuses on ordering, whereas Pearson also reflects relative score differences; high Pearson correlation does not require identical score scales or absolute values.}
    \label{tab:vlm_human_agreement}
\end{table}

Table~\ref{tab:vlm_human_agreement} shows that the trained evaluator agrees with the human preference in 20 of 28 non-tied scenes (71.4\%). Humans prefer X-World~1.6 in 17 of these scenes, giving an always-prefer-1.6 baseline of 60.7\%. The evaluator's Spearman and Pearson correlations with human video penalties are 0.671 and 0.867, respectively. Zero-shot Qwen3-VL-8B-Instruct reaches only 21.4\% pairwise accuracy and 0.097 Spearman correlation under the same inputs and prompts. These results support the value of task-specific training, while the confidence intervals reflect the limited size of the paired set.

At the camera-view level, abnormal-view recall and precision are 71.7\% and 73.0\%. Weighting recall by human penalties and precision by evaluator penalties increases them to 86.0\% and 87.6\%, respectively, indicating better agreement on views carrying larger penalties. These measures characterize the VLM's ability to reproduce the human assessment standard; the benchmark below examines specific consistency properties of the generated videos.

\subsubsection{Benchmark Comparison Protocol}
\label{sec:consistency_protocol}

Checkpoint comparisons use the same scenes, conditioning inputs, camera calibration, resolution, frame rate, rollout duration, and random seeds. Sampling settings are controlled within each comparison and documented if they differ between stages. Evaluator weights and codec settings remain fixed. Scores are aggregated per scene before dataset-level comparison, with equal scene weights and paired changes computed on scenes valid for both checkpoints. Missing tracking observations are marked unavailable rather than assigned a perfect score.

\begin{table}[t]
    \centering
    \footnotesize
    \setlength{\tabcolsep}{3pt}
    \begin{tabular}{lcccccccccc}
        \toprule
        & \multicolumn{3}{c}{Tracking} & \multicolumn{3}{c}{Compression} & \multicolumn{3}{c}{Reconstruction} & Overall \\
        \cmidrule(lr){2-4}\cmidrule(lr){5-7}\cmidrule(lr){8-10}\cmidrule(lr){11-11}
        Model / Variant & App. $\uparrow$ & Decay $\uparrow$ & Ray $\downarrow$ & $\Delta$SSIM $\uparrow$ & $\Delta$PSNR $\uparrow$ & Bits $\downarrow$ & SSIM $\uparrow$ & PSNR $\uparrow$ & MAE $\downarrow$ & Avg. Rank $\downarrow$ \\
        \midrule
        X-World 1.0 & 0.8832 & -0.0257 & 0.1836 & 0.0333 & 0.8587 & 0.6305 & 0.7238 & \underline{21.4834} & \textbf{14.3622} & 3.83 \\
        \midrule
        SF & 0.8856 & -0.0241 & 0.1813 & 0.0336 & 0.8947 & \underline{0.6277} & 0.7279 & 21.2349 & 14.6864 & 3.56 \\
        CSF & \textbf{0.8873} & -0.0234 & \underline{0.1686} & \underline{0.0339} & \underline{0.9183} & 0.6305 & 0.7272 & 21.2381 & 14.7831 & 2.94 \\
        SF+LongTuning & 0.8857 & \underline{-0.0230} & \textbf{0.1680} & 0.0322 & 0.8160 & 0.6445 & \textbf{0.7300} & \textbf{21.5293} & \underline{14.4123} & \underline{2.78} \\
        CSF+LongTuning & \underline{0.8868} & \textbf{-0.0228} & 0.1691 & \textbf{0.0353} & \textbf{1.0039} & \textbf{0.6244} & \underline{0.7286} & 21.3137 & 14.6838 & \textbf{1.89} \\
        \bottomrule
    \end{tabular}
    \caption{\textbf{Consistency comparison with X-World~1.0 and training variants.} SF denotes Self Forcing, CSF denotes Connected Self Forcing, and LongTuning denotes long-horizon tuning. Best and second-best results in each column are shown in bold and underlined, respectively; missing values are excluded. App. denotes Mean Appearance Consistency, Decay denotes Mean Appearance Decay Rate, Ray denotes Mean Ray Residual (m), and Bits denotes the inter/intra encoded-size ratio. A less negative decay slope indicates slower loss of appearance similarity with temporal separation. Both PSNR quantities are measured in dB, and reconstruction MAE uses pixel intensity units (0--255). Avg. Rank is the mean rank across the nine metrics among the five models, with average ranks assigned to ties; lower is better.}
    \label{tab:consistency_stages}
\end{table}

\begin{figure}[!t]
    \centering
    \includegraphics[width=\linewidth]{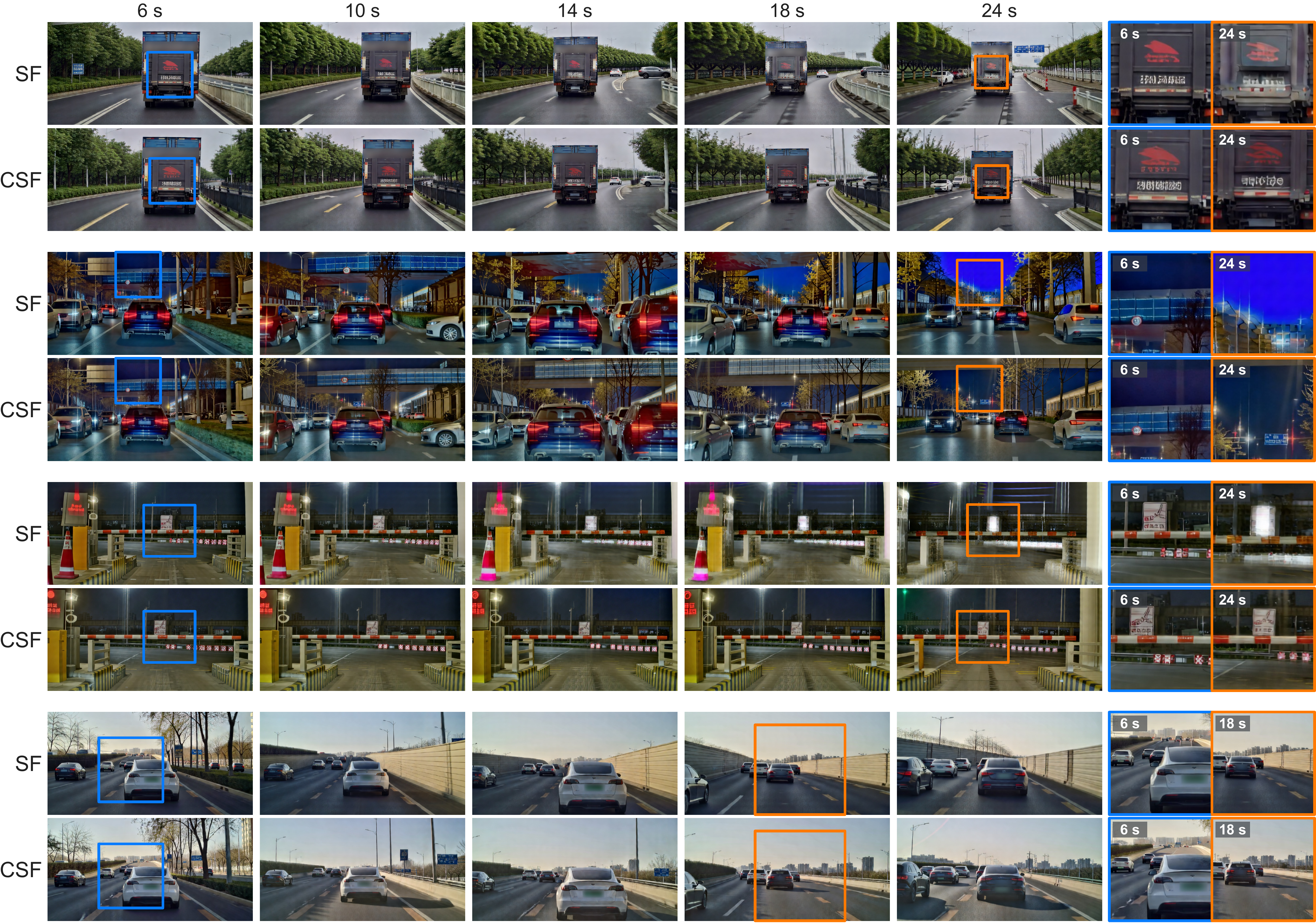}
    \caption{Qualitative comparison of temporal consistency between Self Forcing (SF) and Connected Self Forcing (CSF). Each pair of rows shows synchronized frames from SF (top) and CSF (bottom). From top to bottom, the examples compare vehicle appearance, sky color, gate details, and road structure. Blue and orange boxes mark selected earlier and later regions, respectively, with corresponding enlarged views and timestamps shown in the rightmost column.}
    \label{fig:sf_csf_compare}
\end{figure}

\subsubsection{Results across X-World Variants}
\label{sec:consistency_results}

Table~\ref{tab:consistency_stages} applies the automated benchmark to X-World~1.0~\cite{xworld2026} and four X-World~1.6 training variants: SF, CSF, SF+LongTuning, and CSF+LongTuning. The SF/CSF comparison examines the change from Self Forcing to Connected Self Forcing; comparisons with their long-tuning counterparts examine the effect of extended temporal supervision.

\textbf{Cross-chunk feedback.} Relative to SF, CSF improves mean appearance consistency from 0.8856 to 0.8873 and reduces the cross-view ray residual from 0.1813\,m to 0.1686\,m. Appearance decay and compression quality gains also improve, although the encoded-size ratio increases and reconstruction metrics show mixed changes. These measurements support improved object appearance persistence and cross-view agreement with cross-chunk feedback. Figure~\ref{fig:sf_csf_compare} illustrates the corresponding differences in vehicle appearance, sky color, gate details, and road structure during extended rollouts.

\textbf{Long-horizon tuning.} CSF+LongTuning achieves the slowest appearance decay and the best scores on all three compression metrics. Relative to CSF, it also improves reconstruction, with small regressions in mean appearance consistency and ray residual. The effect depends on the preceding training method: SF+LongTuning achieves the best ray residual and reconstruction scores among the four X-World~1.6 variants, but its compression metrics regress relative to SF. The benchmark thus identifies the specific benefits and trade-offs of extended temporal supervision.

\textbf{Comparison with X-World~1.0.} CSF+LongTuning improves all tracking and compression metrics and reconstruction SSIM over X-World~1.0, while the earlier model retains better reconstruction PSNR and MAE. This comparison supports gains in object consistency and temporal predictability, with a remaining trade-off in pixel-level reconstruction fidelity. One possible contributor is the addition of road-marker and traffic-light control in X-World~1.6: the generator must satisfy additional constraints on marking geometry and traffic-light states while maintaining temporal and cross-view coherence. These requirements may increase generation difficulty and contribute to higher pixel-level reconstruction error under the fixed evaluator.

\section{Applications}
X-World~1.0~\cite{xworld2026} demonstrated closed-loop simulation and online reinforcement learning with control over vehicles, lane markings, and curbs. Version 1.6 improves generation, introduces absolute static scene control to preserve the static environment of a selected location, adds control over traffic-light position and color and over road-surface markings, and expands lane-marking and curb categories. Building on the model-capability evaluation in Section~\ref{sec:eval}, this section describes how Virtual City supports scenario reuse, closed-loop evaluation and training, and target-behavior reproduction in operational driving tasks. We also present dashcam-to-surround completion as an application for synthesizing multi-camera training data from internet videos.

Cloud records allow Virtual City to faithfully reproduce a scene as it originally appeared. Combining these records with X-World~1.6 enables exploration of new situations at the same location, including new traffic flows, different surrounding-vehicle behaviors, and alternative ego actions beyond those in the recording. During this exploration, absolute static scene control keeps the recorded road structure, buildings, and other static surroundings unchanged while the dynamic conditions vary. 

\subsection{Scenario Configuration and Reuse}
A location provides the environmental reference, controls define task conditions, and closed-loop generation enables ongoing policy interaction. Together, they allow the same place to support target-behavior reproduction and policy exploration under different actions. This shared environment connects real-world records to evaluation and training tasks while preserving their scene provenance.

In the proposed platform, a location asset stores scene records and associated information; a scenario configuration specifies the driving task, initial state, and selected conditions; and a generation record links the asset, configuration, generation settings, and multi-camera output. These objects associate multiple tasks with one location and retain input--output relationships for refinement and reuse. Data management, configuration editing, and version associations describe the platform design built on established model capabilities and existing cloud records. Importing additional photographs or street-view material remains a possible extension.

The workflow begins with an existing scene record, an ego starting position, and a driving task. Surrounding-vehicle states or trajectories and relevant lane-marking, curb, road-surface marking, and signal conditions are selected for the task. A left-turn task, for example, can retain the location while varying other vehicles' initial positions and motion to create different encounters. Signal conditions use the currently supported position and color settings. 
% Figure~\ref{fig:vc_scene} illustrates how different interactions share the same static scene reference.

% \begin{figure}[tbp]
%   \centering
%   \includegraphics[width=0.95\linewidth]{figures/virtual_city_scene.pdf}
%   \caption{\textbf{Repeated interactions at one location.} This conceptual illustration shows runs sharing a static scene reference while policy actions and dynamic behavior may differ.}
%   \label{fig:vc_scene}
% \end{figure}

In the proposed workflow, multi-camera outputs generated from a candidate configuration are previewed. Review checks whether fixed structures match the location, vehicles occupy the intended road regions, and visible signal colors and road-surface markings follow the conditions. Views are inspected together, with attention to vehicle motion across cameras and the visibility of key road elements. Where needed, the configuration is revised and regenerated, retaining its conditions and changes from the original.

Selected outputs are linked to their location asset, scenario configuration, and generation settings in this design. Additional recordings extend the scene evidence, while new configurations introduce further traffic tasks at the same location. These associations support inspection and traceability and allow later work to begin from an existing configuration. Reusing a scene provides common experimental conditions; different policy actions can still produce different subsequent observations and trajectories.

\subsection{Closed-Loop Evaluation and Training for End-to-end Policies}
\label{sec:closed-loop-use}
\begin{figure}[htbp]
  \centering
  \includegraphics[width=0.95\linewidth]{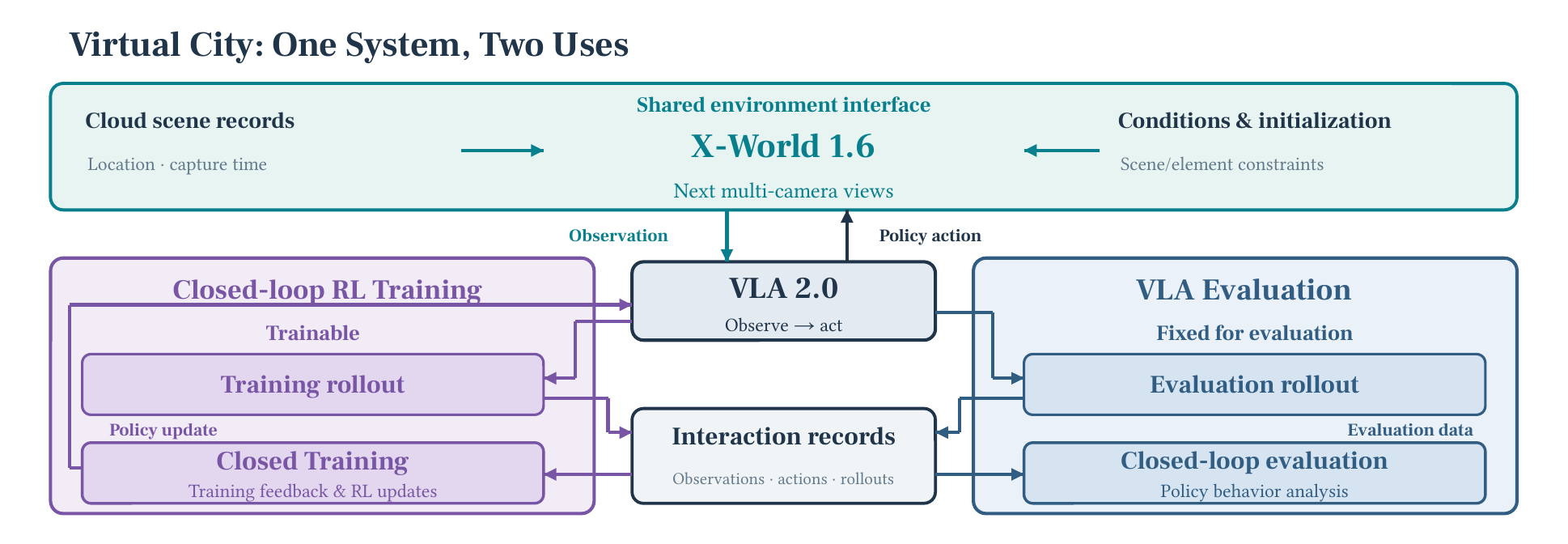}
  \caption{\textbf{Shared interaction pipeline for simulation and RL.} Conceptual illustration of scene-conditioned generation and the policy--environment loop. Interaction records support behavioral evaluation; the RL branch adds training feedback and policy updates.}
  \label{vc_pipeline}
\end{figure}

Simulation and reinforcement learning use the same action--observation interface shown conceptually in Figure~\ref{vc_pipeline}. Initial observations, scene conditions, and the ego state define the interaction starting point. The policy receives synchronized multi-camera observations and produces an action or planned trajectory. The simulator advances the ego state accordingly, and the world model generates subsequent observations using the observation history and scene conditions. These observations return to the policy, completing the interaction loop.

\textbf{Simulation evaluation.} A simulation begins with a recording segment preceding the target event, giving the policy time to perceive, decide, and respond. The interaction then follows the shared pipeline through the specified evaluation window. Cloud records retain the scene reference as the policy's actions change subsequent observations. The resulting interaction is used to examine reproduction of the target failure behavior, supporting diagnosis and regression evaluation.

\textbf{Reinforcement learning.} RoXDrive~\cite{lin2026roxdrive} provides a public research reference for closed-loop policy optimization with a world model. In this methodological setup, the world model remains frozen while the policy is updated using feedback from interaction. Rewards account for safety, progress, comfort, and centering, connecting the collected rollouts to policy updates. These details describe the public research methodology; they are not presented as the configuration of our in-house policy training system.

The X-World 1.6 training workflow has run end to end, and internal road testers have reported qualitative improvements in the resulting policy behavior. Quantitative road-test results and business-specific training details are not reported here.

\subsection{Data Synthesis from Dashcam Videos}
\label{sec:dashcam_data_synthesis}

Internet dashcam videos provide a rich source of driving scenes and traffic interactions, but typically contain only a single forward-facing view. X-World~1.6 converts these clips into synchronized multi-camera videos through dashcam-to-surround completion, producing the camera views required by our production data pipeline. The generated views follow the ego motion and traffic evolution of the source video, extending its recorded sequence to the surrounding views.

This capability broadens the source material available to our data synthesis workflow. Driving scenarios found online can be converted into multi-camera training examples for downstream policies, supplementing our own recordings and helping us meet production data needs.

\subsection{Downstream Utility and Evaluation}
\label{sec:environment-evaluation}
We evaluate how X-World~1.6 supports downstream policy evaluation and failure-case reproduction. We first describe the measures and comparison methods, then report the results and explain their practical value.

\subsubsection{Evaluation Metrics and Methods}

\textbf{Trajectory consistency.} We use final displacement error (FDE) to measure the distance between trajectory endpoints, using trajectory predictions from the corresponding real videos as the reference. Lower FDE means closer agreement with the reference. We report mean FDE and the shares of evaluated cases with FDE below 1\,m, 2\,m, and 5\,m. Each share is the number of cases below the threshold divided by the total number of evaluated cases. The thresholds are cumulative: a case below 1\,m also counts toward the 2\,m and 5\,m thresholds.

\textbf{Target failure reproduction.} We compare the closed-loop results of X-World~1.0 and 1.6 on the same selected test cases. The scene, policy version, task definition, initialization window, and review criteria are kept the same. Reviewers use the original case as the reference and judge how well each simulation reproduces the target failure behavior. Each case is labeled \emph{Improved}, \emph{Comparable}, or \emph{Worse}, depending on whether 1.6 reproduces that behavior more faithfully, similarly, or less faithfully than 1.0. Visual changes are recorded separately; better-looking images alone do not count as better failure reproduction. Each outcome's share uses all reviewed cases as its denominator, and the three shares sum to 100\%.

\textbf{Task-demand summary.} We also report an existing internal score that weights the shares of improved cases in signal-related and road-marking-related task groups by task demand. X-World~1.0 is assigned a reference score of 100, and each demand-weighted percentage point of improvement adds one point. This score counts improvements without subtracting regressions. It summarizes support for selected task needs; it does not measure a net gain or an overall reproduction success rate.

\subsubsection{Results}

Table~\ref{tab:vc_results} compares trajectory consistency and target failure reproduction between X-World~1.6 and 1.0.

\begin{table}[tbp]
  \centering
  \begin{minipage}{\linewidth}
    \small
    \setlength{\tabcolsep}{4pt}
    \renewcommand{\arraystretch}{1.16}

    % Top-aligned panels share one caption; B includes the review interpretation.
    \begin{minipage}[t]{0.40\linewidth}
      \vspace{0pt}
      \textbf{A. Trajectory consistency}\par\nobreak\vspace{3pt}
      \begin{tabular*}{\linewidth}{@{}l@{\extracolsep{\fill}}r@{}}
        \toprule
        Metric & Change \\
        \midrule
        Mean FDE (relative) & $-20.9\%$ \\
        Share with FDE $<1$\,m & $+5.6$\,pp \\
        Share with FDE $<2$\,m & $+9.4$\,pp \\
        Share with FDE $<5$\,m & $+10.6$\,pp \\
        \bottomrule
      \end{tabular*}
    \end{minipage}\hfill
    \begin{minipage}[t]{0.57\linewidth}
      \vspace{0pt}
      \textbf{B. Paired behavior review}\par\nobreak\vspace{3pt}
      \begin{tabular}{@{}>{\raggedright\arraybackslash}p{20mm}>{\raggedleft\arraybackslash}p{13mm}>{\raggedright\arraybackslash}p{\dimexpr\linewidth-33mm-4\tabcolsep\relax}@{}}
        \toprule
        Outcome & Share & Target behavior \\
        \midrule
        Improved & \textbf{19\%} & More faithful reproduction \\
        Comparable & \textbf{77.4\%} & Similar reproduction \\
        Worse & \textbf{3.6\%} & Less faithful reproduction \\
        \midrule
        \multicolumn{3}{@{}l@{}}{\textbf{Improved or comparable: 96.4\%}\quad} \\
        \bottomrule
      \end{tabular}
    \end{minipage}

    \par\vspace{8pt}
    \footnotesize
    \setlength{\parindent}{0pt}
    \setlength{\parskip}{2pt}
    \textit{A.} pp denotes percentage points; FDE-threshold shares are cumulative.\par
    \textit{B.} Shares use all reviewed cases. The summary combines improved and comparable cases; it is not an absolute reproduction success rate. Visual changes alone do not count as behavioral improvement.
  \end{minipage}
  \caption{Trajectory consistency and target-behavior reproduction: X-World 1.6 versus 1.0.}
  \label{tab:vc_results}
\end{table}

\textbf{Trajectory consistency.} X-World~1.6 reduces mean FDE by 20.9\% relative to 1.0. The shares of cases below 1\,m, 2\,m, and 5\,m increase by 5.6, 9.4, and 10.6 percentage points, respectively. These results show closer agreement with the trajectory predictions obtained from real videos.

\textbf{Target failure reproduction.} X-World~1.6 improves reproduction in 19\% of reviewed cases, gives comparable reproduction in 77.4\%, and performs worse in 3.6\%. Together, 96.4\% of cases are improved or comparable. This is a comparison with 1.0, not an absolute reproduction success rate. Comparable cases may still differ in road appearance, signals, markings, or image quality; their target failure behavior is reproduced to a similar degree.

\textbf{Task-demand summary.} Our internal task-demand mapping gives X-World~1.6 a score of approximately 125 against the 1.0 reference of 100. This value has not been recalculated from the current paired-review results in Table~\ref{tab:vc_results}. The task groups may overlap, and the mapping has not been checked task by task. Task volumes, weights, and group details remain confidential, so this supplementary score should be read separately from the direct comparisons above.

\subsubsection{Benefits for Downstream Tasks}

\textbf{Policy evaluation.} Closer agreement with reference-video trajectory predictions supports using generated observations to test policy responses in recorded scenarios. The FDE results provide evidence for this use on the evaluated cases.

\textbf{Failure diagnosis and regression testing.} More faithful failure reproduction helps developers recreate a known problem, inspect how it occurs, and test whether a policy update resolves it. The paired review shows improvements in some cases while preserving comparable reproduction in most others. The cases that become worse identify regressions in the simulator that need further investigation. Reproducing failures more accurately makes the simulator a more useful tool for testing policy changes.

\textbf{Scenario design.} The scene and traffic controls give developers ways to build follow-up tests around a recorded case. Absolute static scene control preserves the recorded road layout, buildings, and background, while road-marker and traffic-light controls provide directional cues and signal states for tasks such as stopping, proceeding, and turning. These controls support testing different traffic conditions at the same location. The reported comparisons assess X-World as a complete system and do not isolate the benefit of any individual control. The results apply to the evaluated cases and are not extrapolated to the full simulation collection.

\section{Conclusion and Limitations}
\label{sec:conclusion}

X-World~1.6 strengthens scene control, long-horizon interaction, and model evaluation on top of an established driving simulator. Absolute static scene control connects recorded locations to configurable generation, while improved post-training and evaluation support more reliable use in the shared closed-loop simulation and policy-training workflow. Together, these advances support the Virtual City product concept.

Two limitations remain central. Although the model supports multi-view rollouts exceeding 30 seconds, prolonged occlusions and long-range movement can still prevent previously observed scene content from being recovered faithfully. Long rollout duration therefore does not guarantee persistent scene memory. In addition, temporally inconsistent control inputs can produce discontinuities or inconsistent behavior. Improving scene memory and reconciling imperfect controls with coherent world evolution are priorities for further post-training and model development. The reported consistency and application results should be interpreted within these limits and the coverage of the supplied scene references.

\section{Contributors}
\label{sec:contributors}

The following individuals contributed to the development of X-World 1.6.
Names are listed alphabetically by given name within each group.

\medskip
\noindent\textbf{Advisors}\\
Xianming Liu, Yu Zhang

\medskip
\noindent\textbf{Project Lead}\\
Boyang Wang

\medskip
\noindent\textbf{Pre-training}\\
Jinhao Deng, Kangjie Chen, Ruixin Liu, Xiangyu Li, Xuejie Lv, Yuqi Zhang

\medskip
\noindent\textbf{Post-training}\\
Chaoda Zheng, Dongbin Zhang, Guangfeng Jiang, Hanlong Chen

\medskip
\noindent\textbf{Simulation and Closed-Loop Policy Training}\\
Kai Yang, Linkun Xu, Shijia Chen, Shuang Deng, Sijia Chen,
Xingjie Liu, Xuezhi Yang, Yuxin Peng

\medskip
\noindent\textbf{Infrastructure}\\
Jianlei Zheng, Yixiao Zeng, Yuecheng Ma

\medskip
\noindent\textbf{Data}\\
Rongliang Cheng, Tenglong Gu, Tianyuan Xie, Xinlong Zheng, Yi Guo

\medskip
\noindent\textbf{Additional Contributors}\\
Boyi Sun, Hongbin Lin, Minqi Wang, Puyi Wang, Sin Wai Choo, Yuhao Wang

\clearpage
\appendix
\section{Base-Model Implementation Details}
\label{app:method}
\subsection{GeoPE Descriptor and Fusion}
Figure~\ref{fig:geope} summarizes the three steps of GeoPE: describing a patch by its viewing ray, preserving the geometry of the source frames represented by a video latent, and adding the resulting embedding to the visual token. The same construction applies to reference observations and noisy target tokens.

\begin{fixedfigure}
\centering
\includegraphics[width=\linewidth]{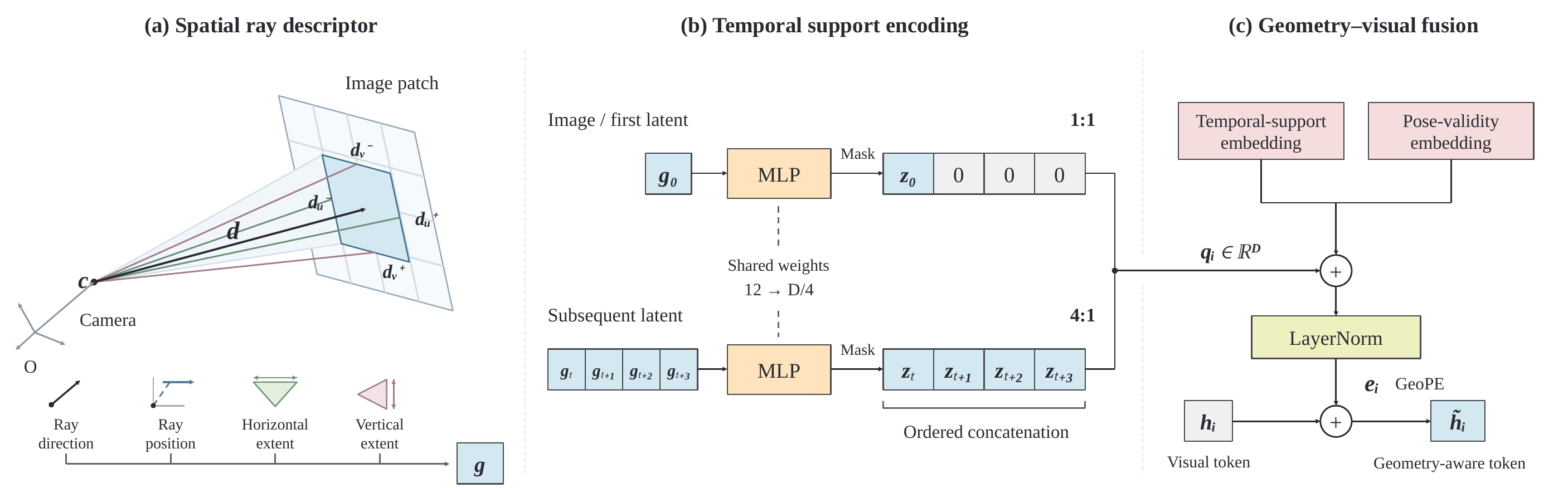}
\caption{\textbf{Geometry-aware positional encoding.} (a) Ray direction, position, and patch extent form the spatial descriptor. (b) A shared MLP encodes one or four source-frame descriptors, which occupy ordered slots under the corresponding temporal-support mask. (c) Temporal-support and pose-validity embeddings are fused with the geometry, normalized, and added to the visual token.}
\label{fig:geope}
\end{fixedfigure}

In panel (a), for camera center $\mathbf c$ and patch-center unit direction $\mathbf d$, GeoPE uses
\begin{equation}
\mathbf g=[\mathbf d;\operatorname{symlog}(\mathbf c\times\mathbf d);
\Delta\mathbf d_u;\Delta\mathbf d_v]\in\mathbb R^{12}.
\label{eq:geope_geometry}
\end{equation}
The direction and Pl\"ucker moment locate the viewing line. The transform $\operatorname{symlog}(x)=\operatorname{sign}(x)\log(1+|x|)$ compresses the moment's dynamic range. Angular extents use $\Delta\mathbf d_a=(\mathbf d_a^+-\mathbf d_a^-)/2$, for $a\in\{u,v\}$, with unit directions at opposite patch-boundary midpoints after calibration and distortion correction.

Panel (b) uses four ordered slots to retain frame-specific geometry under temporal compression. Independent images and the first video latent activate only the first slot; subsequent latents use four consecutive descriptors. A shared two-layer SiLU MLP $f_\theta:\mathbb R^{12}\rightarrow\mathbb R^{D/4}$ embeds each slot; panel (c) fuses the resulting embeddings as follows:
\begin{equation}
\begin{aligned}
\mathbf e_i&=\operatorname{LN}\!\left(
\mathop{\Vert}_{k=1}^{4}[m_{ik}f_\theta(\mathbf g_{ik})]
+E_{\mathrm{support}}(s_i)+E_{\mathrm{pose}}(v_i)\right),\\
\widetilde{\mathbf h}_i&=\mathbf h_i+\mathbf e_i.
\end{aligned}
\label{eq:geope_fusion}
\end{equation}
The mask $m_{ik}$ removes inactive or pose-invalid slots after the MLP. Learned $D$-dimensional embeddings encode one- versus four-frame support $s_i$ and pose validity $v_i$, retaining these distinctions even when geometric slots are masked. The same encoder is used for reference and noisy target tokens.

\subsection{Pose and Signal Conventions}
\label{app:pose_signal}
For the 3DGS conditioning described in Section~\ref{sec:absolute_static_control}, the per-frame relative pose is
\begin{equation}
\boldsymbol{\Delta}_t
=\mathbf M_0(\mathbf C_t^{\mathrm{gt}})^{-1}
\mathbf C_t^{\mathrm{render}}\mathbf M_0^{-1},
\label{eq:pose_delta}
\end{equation}
where $\mathbf C_t^{\mathrm{gt}}$ and $\mathbf C_t^{\mathrm{render}}$ are the target and reference camera poses, and $\mathbf M_0$ is the inverse of the fixed ego-to-camera extrinsic transformation. We decompose this relative pose into rotation $\mathbf R_t\in\mathrm{SO}(3)$ and translation $\mathbf T_t\in\mathbb R^3$, and encode their normalized pose parameters using Fourier features and an MLP. The resulting pose embedding is added to the reference GeoPE embeddings and broadcast across the patches of each latent frame. The same relative pose aligns the reference geometry with the target scene frame. The reference view matrix used by GeoPE is
\begin{equation}
\mathbf W_{\mathrm{ref}}(t)=\mathbf M_0^{-1}\boldsymbol\Delta_t^{-1}.
\label{eq:patch_cape_delta}
\end{equation}

Traffic-light conditioning follows a different first-latent convention from GeoPE: its initial state is repeated across four source-frame head groups. Subsequent latents use four successive source-frame states. Signal masks restrict direct branch updates to projected light regions, while shared agent, lane, and marker conditions use token-validity masks in their post-fusion cross-attention branch.

\section{Evaluation Definitions and Protocol}
\label{app:evaluation}

This appendix specifies the VLM reward and the custom consistency diagnostics used in Section~\ref{sec:eval}. All evaluators are fixed during checkpoint comparisons.

\subsection{VLM Reward}
\label{app:vlm_reward}

The evaluator predicts anomaly status, categories, frame intervals, and boxes normalized to $[0,1000]$. GRPO uses
\begin{equation}
R = w_1 R_{\mathrm{fmt}} + w_2 R_{\mathrm{stat}} + m \left( w_3 R_{\mathrm{type}} + w_4 R_{\mathrm{temp}} + w_5 R_{\mathrm{spa}} \right),
\label{eq:vlm_reward}
\end{equation}
The terms reward format validity, correct status, category overlap, temporal IoU, and mean spatial box IoU on frames with both predictions and annotations. The indicator $m$ equals 1 for abnormal samples and 0 otherwise; localization terms receive the largest weights.

\subsection{Tracking Diagnostics}
\label{app:tracking_metrics}

For normalized masked-object embeddings $\mathbf e_{k,t}$, compare valid same-track pairs at intervals $\mathcal D=\{1,2,4,8,16,32\}$ frames. A continuity gate removes unreliable associations; missing observations are not interpolated. With $\mathcal P$ the valid triples,
\begin{equation}
    C_{\mathrm{app}}=
    \frac{1}{|\mathcal P|}\sum_{(k,t,\delta)\in\mathcal P}
    \mathbf e_{k,t}^{\top}\mathbf e_{k,t+\delta}.
    \label{eq:wm_appearance_consistency}
\end{equation}
To distinguish similar average scores with different temporal behavior, we also measure how similarity changes as the frame interval increases. For each track, let $\bar c_k(\delta)$ be its mean similarity at interval $\delta$, and let $\mathcal D_k$ contain the intervals with valid samples. We fit
\begin{equation}
    (a_k,b_k)=\underset{a,b}{\operatorname{arg\,min}}
    \sum_{\delta\in\mathcal D_k}
    \left[\bar c_k(\delta)-a-b\log_2\delta\right]^2,
    \qquad
    D_{\mathrm{app}}=\frac{1}{|\mathcal K|}\sum_{k\in\mathcal K}b_k,
    \label{eq:wm_appearance_decay}
\end{equation}
Here $\mathcal D_k$ contains valid intervals and $\mathcal K$ tracks with at least two distinct intervals. Mean Appearance Decay Rate $D_{\mathrm{app}}$ describes the change in similarity when the interval doubles. A slope closer to zero indicates slower decay.

For matched same-category cross-view observations, let $\mathbf o_a,\mathbf o_b$ be camera centers in the common ego frame and $\mathbf d_a,\mathbf d_b$ the unit rays through mask centroids:
\begin{equation}
    r_p=\min_{s,u\in\mathbb R}
    \left\|\mathbf o_a+s\mathbf d_a-\mathbf o_b-u\mathbf d_b\right\|_2,
    \qquad
    R_{\mathrm{ray}}=\frac{1}{|\mathcal Q|}\sum_{p\in\mathcal Q}r_p,
    \label{eq:wm_ray_residual}
\end{equation}
The set $\mathcal Q$ contains same-frame associations with non-parallel rays and closest points in front of both cameras. Residuals are in meters. Because matching itself uses geometric compatibility, interpretation should consider valid cross-view coverage.

\subsection{Compression Diagnostics}
\label{app:compression_metrics}

We encode each video with libx264 at CRF~36, using an inter-frame GOP of~128 and an all-intra counterpart; adaptive scene-cut insertion is disabled. Let $S$ and $P$ denote decoded-video SSIM and PSNR against the input, computed on luminance, and $B$ denote file size:
\begin{equation}
    \Delta\mathrm{SSIM}=S_{\mathrm{inter}}-S_{\mathrm{intra}},
    \qquad
    \Delta\mathrm{PSNR}=P_{\mathrm{inter}}-P_{\mathrm{intra}},
    \qquad
    R_{\mathrm{bits}}=\frac{B_{\mathrm{inter}}}{B_{\mathrm{intra}}}.
    \label{eq:wm_compression_consistency}
\end{equation}
Larger quality gains and a lower encoded-size ratio indicate greater benefit from inter-frame prediction. The two streams have equal duration, so the size ratio also equals their average bitrate ratio. The intra baseline reduces sensitivity to spatial complexity but does not remove dependence on motion or content.

\begin{figure}[!htbp]
    \centering
    \includegraphics[width=\linewidth]{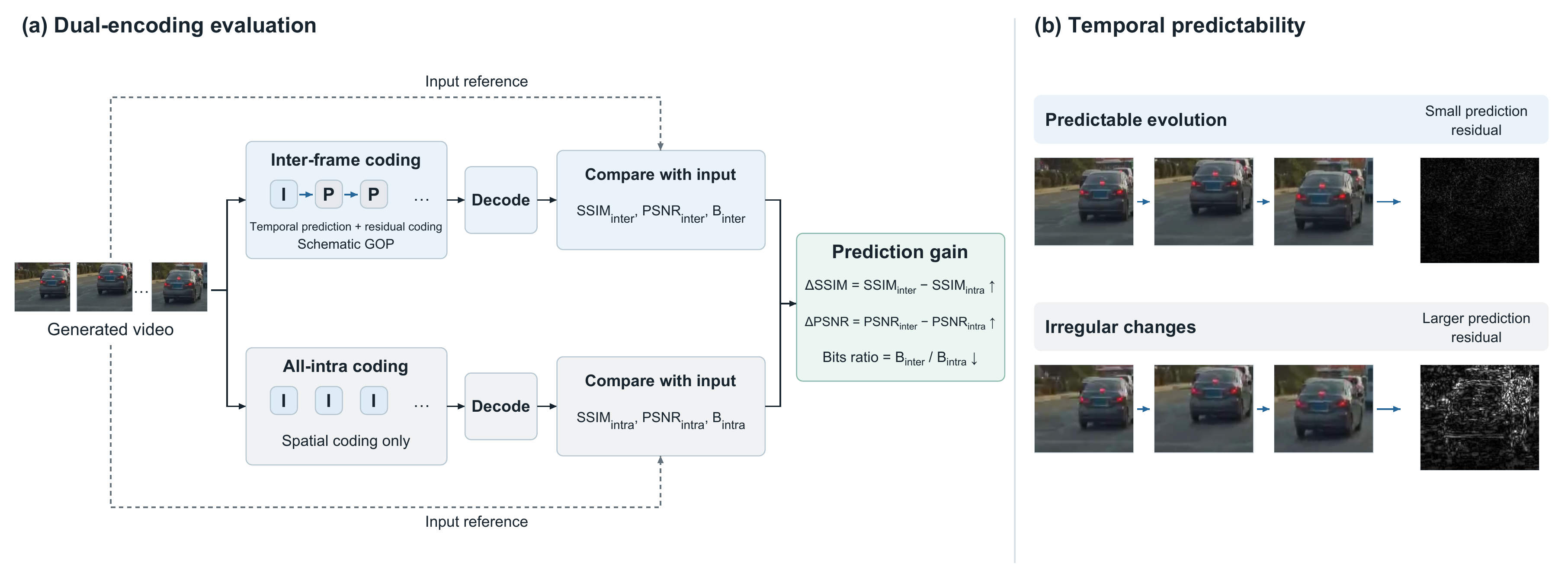}
    \caption{\textbf{Compression-based temporal consistency.} (a) Inter-frame and all-intra coding of the same video provide quality gains and an encoded-size ratio. (b) Conceptual illustration of larger prediction residuals from irregular appearance changes under the same motion.}
    \label{fig:compression_consistency}
\end{figure}

\subsection{Reconstruction and Aggregation}
\label{app:reconstruction_metrics}

UniSplat processes synchronized views jointly at each time step and retains Gaussian history across time:
\begin{equation}
    (\mathcal G_t,h_t)=F_{\mathrm{rec}}(\{I_t^v\}_{v=1}^{V},\mathcal C,h_{t-1}),
    \qquad
    \hat I_t^v=\mathcal R(\mathcal G_t;\mathcal C_v),
    \label{eq:wm_reconstruction}
\end{equation}
Here $\mathcal C$ is calibration, $h_t$ temporal memory, $\mathcal G_t$ the reconstructed Gaussian representation, and $\mathcal R$ the renderer. Camera poses are fixed without refinement. Compare rendered and input RGB frames at reconstruction resolution in the 8-bit pixel domain using SSIM, PSNR, and MAE. This tests reconstructability at observed views, not held-out novel-view synthesis.

Compression and reconstruction scores are averaged across views within each scene. Tracking aggregates valid observations as defined above. Dataset comparisons weight scenes equally and use the scenes valid for both checkpoints; missing-track scenes remain unavailable and their counts should accompany the results. Fixed scene sets, conditions, calibration, resolution, frame rate, duration, seeds, evaluator weights, and codec settings support paired comparison. Any stage-dependent sampling differences must be documented.

\section{Training and Inference Implementation Notes}
\label{app:systems}

\subsection{Packed Causal Evaluation}

For one candidate suffix $Z=(z_b,\ldots,z_E)$ after clean prefix $P$, each target block is independently noised as $x_j=(1-\sigma_t)z_j+\sigma_t\epsilon$. Current, old, and reference models build separate detached KV banks from the clean trajectory. Clean-bank blocks themselves attend only to their permitted past and their own block, preventing future leakage through cached representations. Packed query block $j$ accesses
\begin{equation}
\mathcal S_j
=
\mathcal W_j\!\left(
\operatorname{Tok}(P\mathbin{\Vert}z_{b:j-1})
\cup
\operatorname{Tok}(x_j)
\right),
\label{eq:block_visibility}
\end{equation}
where $\mathcal W_j$ enforces the original sink/window constraints and $z_{b:j-1}$ is empty for $j=b$. Historical keys and values come from the clean bank; the current query supplies its own noisy keys and values. Its clean target, future clean blocks, and other noisy blocks remain masked. Building the clean bank first replaces $H=E-b+1$ sequential target calls with one packed call per model role and candidate batch, without backpropagating through trajectory generation.

\begin{figure}[t]
    \centering
    \includegraphics[width=0.98\linewidth]{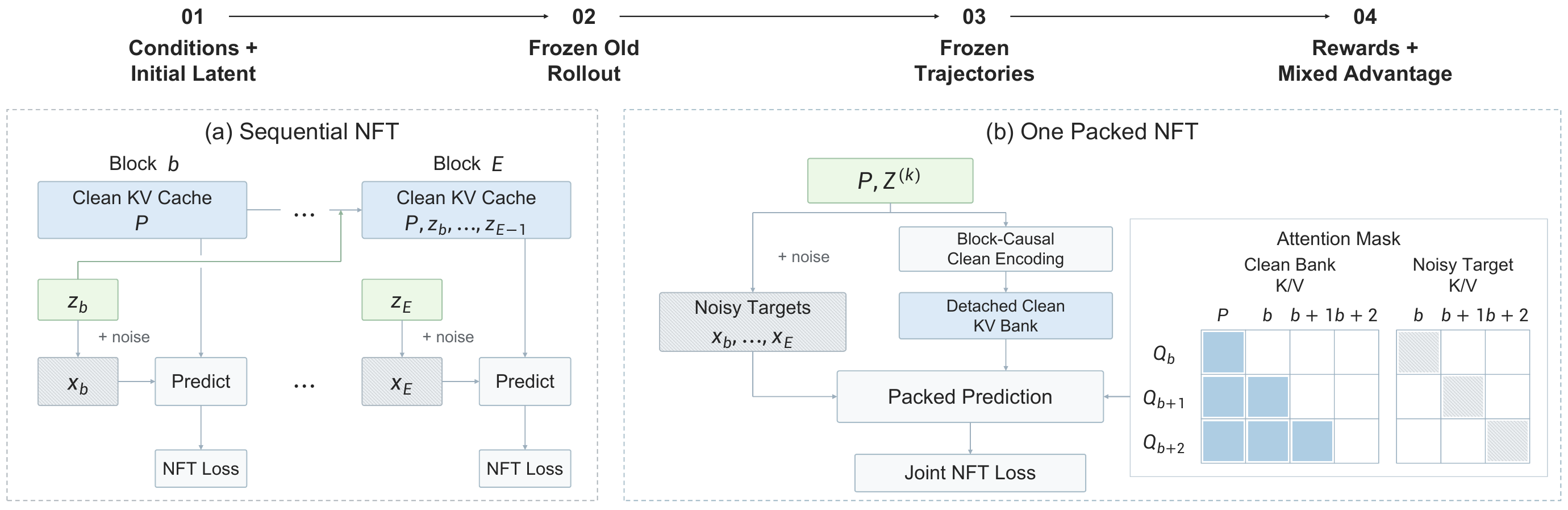}
    \caption{\textbf{One Packed NFT.} Sequential evaluation interleaves target prediction and clean-cache extension; packed evaluation uses a detached clean KV bank while preserving the same causal visibility. Both use frozen old-model trajectories.}
    \label{fig:one_packed_nft}
\end{figure}

\subsection{Kernel and Decoder Execution}
\label{sec:inference_dit}
\label{sec:inference_vae}

The BF16 DiT uses the SAIL implementation of FlashAttention-3~\cite{shah2024flashattention3}. Fused kernels combine normalization, modulation, gated residual updates, and Q/K rotary processing to reduce launches and intermediate tensors. The VAE retains \texttt{NDHWC} activations between convolutional layers; fused input kernels combine causal padding, cached-history insertion, normalization, and activation. Dominant $3\times3\times3$ convolutions use AIU implicit GEMM without an explicit \texttt{im2col} buffer.

For each camera, decoding starts from the encoded last frame of the preceding four-frame interval. The decoder snapshots causal convolution states after this boundary prefix. Later requests in the same interval resume from that snapshot and decode only the temporal prefix needed for the requested frame; the fourth frame is re-encoded to establish the next boundary.

\subsection{Session State and Numerical Semantics}
\label{sec:inference_serving}

Initialization, denoising, and decoding run in separate worker processes. Sessions use preallocated caches, dedicated streams, and pinned transfer buffers. Each session keeps one denoiser and selects a decoder on its first decode request. X-Cache residuals are held in pinned host memory and transferred asynchronously when reused, enabling three sessions per denoiser on the reported four-PPU node.

Exact kernel optimizations preserve the reference reduction order and intermediate BF16 rounding boundaries, including FP32 convolution accumulation before round-to-nearest-even conversion. Deterministic multi-stream BLAS workspaces and dedicated hardware queues prevent interleaving from changing these semantics. This preserves a clear distinction between implementation changes that are numerically exact and X-Cache's deliberate residual approximation.

\clearpage
\bibliographystyle{plainnat}
\bibliography{main}

\end{document}